\documentclass[runningheads]{llncs}
\usepackage{graphicx}
\usepackage{algorithm}
\usepackage{pseudo}
\usepackage{subfigure}
\usepackage{orcidlink}
\newtheorem{axiom}{Axiom}
\newtheorem{define}{Definition}

\begin{document}

\title{What is Meant by AGI? On the Definition of Artificial General Intelligence}
%
%
\author{Bowen Xu \orcidlink{0000-0002-9475-9434}}
%
\authorrunning{B. Xu}
%
\institute{Department of Computer and Information Sciences, \\
Temple University, Philadelphia, USA\\
\email{bowen.xu@temple.edu}}
\maketitle              
\begin{abstract} 

This paper aims to establish a consensus on AGI's definition. General intelligence refers to the adaptation to open environments according to certain principles using limited resources. It emphasizes that \textit{adaptation} or \textit{learning} is an indispensable property of intelligence, and places the controversial part within the principles of intelligence, which can be described from different perspectives. 

\end{abstract}
\section{Introduction}


Why do we take this philosophical question so seriously? It is neither because broad interests in Artificial General Intelligence (AGI) have raised~\cite{mitchell2024debates} over the years, nor because of any endeavors for the right to use the term AGI. Rather, \textit{what intelligence means} and \textit{what AGI means} are the most fundamental questions of AGI research, though these two questions are highly related. Almost all questions and debates on AGI can be guided to asking the definition of AGI, such as ``is AGI possible to realize?'', ``how to realize AGI?'', ``is that thing an AGI system?'', ``how to evaluate AGI?'', ``whether will AGI save humanity or destroy it?'', and so forth. Leaving the plentiful questions aside, in this paper, we focus on the question ``what is meant by AGI?'', aiming to give a clear definition, and attempting to end the lack of specification of the term ``AGI''. 

In the past, researchers did not make a clear definition of AGI, but it does not mean that AGI can be defined in arbitrary ways. There are some previous descriptions of AGI:
\begin{enumerate}
    \item AGI is the initial and ultimate goal of AI research;
    \item ``AGI'' refers to the original ``AI'' at the inception of the field of artificial intelligence;
    \item AGI systems should not be designed for specific problems, but they are general-purposed systems without designers' or developers' specifying problems to be solved by the systems;
    \item AGI systems are general systems instead of general algorithms, meaning that after deployed, no human developers need to intervene in the source code of the system, by contrast, an algorithm is general in the sense that human developers can apply it to various problems by generating a solver instance for each problem;
\end{enumerate}
and so on. The meaning of AGI will not be completely clear until a system comes into being and is accepted by almost all researchers to count as AGI. 

As claimed in the proceedings of the AGI Workshop in 2006~\cite{goertzel2007agi}, ``we believe that at the current stage, it is too early to conclude with any scientific definiteness which conception of `intelligence' is the `correct' one.'' Nevertheless, 18 years have passed, should we make it more clear the meaning of the term we use and try to find the (minimal) consensus of the community? I believe it is the time now. This paper endeavors to summarize the minimal consensus of the community, consequently providing a justifiable definition of AGI. It is made clear what is known and what is controversial and remains for research, so as to minimize the ambiguous usages as much as possible in future discussions and debates.

\section{Basis}

There is no doubt that ``intelligence'' is a fat concept. Different characteristics of the human mind were come up with by direct observations. The major perspectives~\cite{legg2007defs}\cite{wang2019defining} are (1) ``intelligence is something emergent from the brain,'' (2) ``intelligence manifests itself through sophisticated (especially, human-like) behaviors,'' (3) ``intelligence is the capability to solve problems,'' (4) ``intelligence is the summation of various cognitive functions,'' and (5) ``intelligence is the capability to adapt to environments''; sometimes intelligence refers to a measure of how smart a subject is~(\textit{e.g.}, \cite{chollet2019measure}). 
Generally speaking, they are all correct and no such one is more correct than the others, because they directly stem from observations, and they actually refer to seemingly similar but essentially different concepts while using the same word ``intelligence'' to refer to them. However, in the context of AGI, which is the better understanding of the term ``intelligence''?

Except for human intelligence, various other forms of intelligence also make sense, for example, animal intelligence, alien intelligence, and so on. In this sense,  ``intelligence'' at the most abstract level may be a more suitable one. In addition, to say different problems necessitate different kinds of ``intelligence'' is no better than to say solutions for different problems stem from something invariant that is called ``intelligence''. That is why the learning capability is more important than other capabilities for specific problems.

Therefore, it is better to define intelligence as a meta-capability -- or learning capability, 
adaptability, whatever you name it. Few researchers might neglect that learning is a necessary aspect of intelligence, or at least a critical one, but sometimes they overlooked this meta-capability while focusing too much on the complexity of the problems to be solved. After a problem was solved, people looked back and doubted whether it was the ``real'' intelligence (\textit{a.k.a.} the ``AI effect''~\cite{geist2016aieffect}). They found that the problem is actually solved by human developers rather than machines themselves. This is because machines solve problems without a procedure of acquiring problems-related knowledge.

Standing on a new starting point where we have had the experience and lessons from our predecessors, the following \textit{axiom} is adopted as the basis of this paper:

\begin{axiom}\label{ax:learn}
For anything that can be viewed as an information system, \textit{if it is intelligent, it can learn, that is, it can adapt to its environment}.
\end{axiom}

In other words, \textit{if a system cannot learn, it is not intelligent}. This axiom is sufficiently intuitive to many people.
~\footnote{However, a few people might argue that even a calculator is intelligent, though it is a different aspect from human intelligence: it surpasses humans in the calculation capability which was the the unique activity of the human mind. However, I argue that different problem-solving capabilities can be viewed as \textit{skills} that are acquired through \textit{intelligence}.}
If one accepts Axiom~\ref{ax:learn},  some implications are directly derived: for example, a human may be intelligent, a calculator is not intelligent at all, a microbe may be intelligent if it can learn; a computer may be intelligent if it can learn; and so forth.

Here, the definition of adaptation should be given:

\begin{define}
For any information system, to adapt to its environment means that (1) externally, when the environment is relatively stable, evaluated by certain metrics, the performance of the system tends to become better, and (2) internally, the internal status of the system changes towards certain directions.
\end{define}
The former illustrates the expectation of the system's observable behaviors, while the latter corresponds to principles inside the system itself. The environment may be continuously changing, thus a system cannot be performant if future situations are not consistent with its past experience at all. The system has goals to achieve, thus the change of the internal states is not fully arbitrary. 

Another self-evident statement is that a system cannot work with infinite resources, especially for an intelligent system. In A.M. Turing's perspective on computer, he assumed theoretically that a computer can have infinite memory and extremely fast computing speed~\cite{turing1950}. However, human beings in the real world always act with highly limited memory and limited execution speed. We should view this as a theoretical constraint if we are creating ``real'' intelligent machines. Even for Large Language Models \cite{bubeck2023sparks-agi}, the resources are still limited, both in theory and in practice. Discarding this constraint will lead researchers to omit an important issue of intelligence -- the forgetting procedure. Facing the real world, resources are often insufficient; even when solving specific problems, when data volume becomes large and exceeds the system's capacity, it has to forget something rationally.
Consequently, the following axiom is claimed explicitly:

\begin{axiom}
For any information system that is intelligent, both in practice and in theory, it has limited computational resources, including memory (that is, spatial resource) and information processing speed (that is, temporal resource).
\end{axiom}

The two statements are too intuitive to be worth further arguments, so I call them axioms instead of assumptions. However, not all researchers follow them as their theories' constraints. For example, in the early stage of AI, people sought the \textit{general problem solver} and invented exhaustive search algorithms (\textit{e.g.}, Breadth-First Search), which are now mostly counted as a part of computer science. In the future, when ``real AIs'' occur, they will interact with the world and human beings by themselves without the designers' intervention; they may become humans' friends; they may help humans to independently explore the unknown environments on remote planets. I believe readers who have the vision of ``real AIs'' will readily accept these two axioms.

Cognitive scientists may argue why not to claim more axioms considering perception, reasoning, planning, and so on. Computer scientists may argue that solving complex problems is also necessary, and some general strategies could be a part of intelligence. Although there are much less disagreements on the learning capability, we can see the disagreements with each other in plentiful definitions of intelligence~\cite{legg2007defs}. I adopt a compromised solution, putting them into the controversial part of the definitions (see the following sections).

\section{Definitions}

I do not wish to give the impression that I completely reject previous AI research, but rather I wish to find a definition of intelligence compatible with previous AI research, especially that in Machine Learning (ML). In the meanwhile, I suggest to make a clear discrimination by defining another concept, \textit{general intelligence}.

\subsection{On ``Intelligence''}

A well-accepted definition~\cite{monett2018survey}, though which many people also object to~\cite{monett2020wang-def}, of intelligence is Pei Wang's~\cite{wang2019defining}: Intelligence is the capacity of an information-processing system to \textit{adapt} to its environment while operating \textit{with insufficient knowledge and resources}. This definition has grasped some critical aspects of intelligence and is a representative one. Actually, I argue that Wang's definition is almost equivalent to the definition of ``general intelligence'' in this paper (see Sec.~\ref{sec:general} and Sec.~\ref{sec:comparisons}). Here, a similar but more lenient one is proposed:

\begin{define} \label{def:intelligence}
From one perspective, intelligence is the capability for an information system to adapt to the environment with limited computational resources. From another perspective, intelligence is a collection of principles.
\end{define}

For ease of description in the following, I denote the collection of principles as $\mathcal{P}$.

The key difference here from Wang's definition is that the computational resources are not insufficient but limited. That is to say, the traditional Machine Learning methods could still be viewed as intelligent.

One may argue that ``limited resources'' is too trivial to be announced in the definition, since in practice, if a system's resources are insufficient, we just need to expand the resources to meet the demand of the algorithms we use. 

I agree with this objection if we only consider special-purpose algorithms. However, the story is fully different when it comes to general-purpose systems. I put it into Def.~\ref{def:intelligence} just for three reasons -- (1) that it is not wrong,  (2) to keep consistent with Def.~\ref{def:gi} (general intelligence), as well as (3) to highlight this realistic constraint and its potential theoretical outcomes (\textit{e.g.}, the forgetting issue mentioned above).

According to Def.~\ref{def:intelligence}, a neural network governed by the gradient descent mechanism is intelligent (if we suppose $\mathcal{P}$ is empty), since it can learn with limited resources; but in contrast to a few people's view, a brute-force search algorithm is not intelligent at all for it unable to learn, though it is highly skilled at solving complex problems.

Intuitively, if an organism exhibits adaptive behaviors, even though it is not as complex as human beings, we would still expect it to be somewhat smart or intelligent.
In this definition, \textit{adaptation} is a necessary condition of intelligence. This is also implied by Axiom~\ref{ax:learn}.

If a brain-like structure does not involve \textit{learning}, but rather merely strictly repeats its responses to stimuli, it is not intelligent (recall the calculator example). In this view, $\mathcal{P}$ is described via the neuroscience language.

For a human-behavior imitator, the behaviors of adaptation (or learning) are considered non-negligible. Without this, a machine cannot play satisfactorily the Imitation Game proposed by Turing~\cite{turing1950} (\textit{a.k.a.} the Turing Test).
In this view, $\mathcal{P}$ is empty.

A lot of AGI projects focus on \textit{cognitive architectures}~\cite{kotseruba2020cogarch}. They emphasize cognitive functions, such as perception, reasoning, planning, decision-making, \textit{etc}. This does not violate the definition above, since they see intelligence from another perspective -- what the principles of intelligence are. They propose a collection of principles and describe how they relate to each other. In this view, $\mathcal{P}$ is described via the cognitive science language.

There are also some worthwhile principles implied by problem-solving algorithms, for example, Monte-Carlo Tree Search, one of the core techniques in AlphaGo~\cite{silver2016alphago}, may be involved in a certain principle describing how an intelligent system seeks potential knowledge to use. In this view, $\mathcal{P}$ is described via the computer science language.

Definition~\ref{def:intelligence} is compatible with the intuition that ``intelligence is the capability of solving complex problems'', but the latter does not imply the former. A machine can solve a problem in two ways: (1) humans directly tell it how (by designing algorithms and programming) and it executes the algorithms; (2) the machine tries to solve the problem  (via discovering algorithms by itself) without human designers directly modifying its code. The key difference between the two ways is whether there is an adaptation procedure. Typical ML systems work in the second way, however, we can say that typical ML systems exhibit intelligence in the training stage but no intelligence in the test stage. 















\subsection{On ``Artificial''}

Unlike intelligence that exists in nature, Artificial Intelligence (AI) should be achieved by artificial means. However, if one thinks about it carefully, there are difficulties with the definition of ``artificial'', especially when it comes to bio-technology. There are at least four possible ways that AI can be implemented through --

\begin{itemize}
    \item \textbf{classical computer with Von Neumann's architecture}: the meaning of a classical computer here is relative to the quantum computer that will be mentioned below,
    \item \textbf{heterogeneous classical computers}: the algorithm is accelerated by GPU, FPGA, analog circuit, \textit{etc.}, so that the time complexity of some calculations is reduced, and even becomes a small constant, 
    \item \textbf{quantum computer}. Certain computational processes that are extremely complex in classical computers, accelerated by quantum computing, may become acceptable as part of intelligent processes, and
    \item \textbf{biological computers}. Biological computing may extend computing power further, thereby breaking some of the constraints of computational complexity. For example, DNA storage, parallel computation of neurons, \textit{etc.}, can enable complex calculations to be completed in a very short duration.
\end{itemize} 

The problem is with bio-computers. Imagine, if one day, people can produce life in a test tube through biotechnology, and the organism is as intelligent as human beings; whether is its intelligence ``artificial intelligence''? One can design a biological computer in which intelligent programs can run; is such an agent ``artificial intelligence''? 
Intuitively, we might assume that the latter's intelligence is artificial and the former's is not. 

How can a biological computer equipped with intelligent programs be distinguished from intelligent life produced in a test tube? One possible view is that the intelligence of the former is artificial and that of the latter is not, because the intelligence of the latter is given by humans, while the intelligence of the former is spontaneously emergent. However, if we consider the intelligence generated by ``artificial neural networks'' to be artificial, then many systems, including large models of artificial neural networks, have emerging capabilities that humans have not anticipated, and it is as difficult for humans to understand and explain as the intelligence emerging in test tubes. So, why is the intelligence of an artificial neural network ``artificial'', but what emerges from a test tube is not? The concept of ``artificial'' has become blurred here.

To avoid making the issue too complex, we just take the convention that clone technology is not under consideration in the context of AI.

\subsection{On ``General''}~\label{sec:general}

I define \textit{general intelligence} and the corresponding arguments in this section.
Some researchers understand ``general'' as ``capable of solving a wide range of tasks or problems'', and some other researchers see it as ``capable of solving problems that are not predetermined or realized by developers''. They are both correct, and I argue that both of them are necessary but insufficient for \textit{general intelligence}. Let us look at the definitions first

\begin{define} \label{def:gi}
From one perspective, intelligence is the capability for an information system to adapt to the \underline{open} environment with limited computational resources. From another perspective, intelligence is a collection of principles.
\end{define}
Compared to Def.~\ref{def:intelligence}, the additional condition in Def.~\ref{def:gi} is that the environment is ``open''. Another difference is that the collection of principles here, denoted as $\mathcal{P}_G$, is a superset of $\mathcal{P}$. 
In Def.~\ref{def:gi}, by ``open environment'', we mean that subjects' activities are confined to a relatively limited area during their lifetime, beyond which the circumstances are unknown to the subject. Consequently, the environment the subject faces may change (even fundamentally), and the future may not be consistent with past experiences; the regularities previously recognized by a subject might be overturned. Additionally, the term ``open environment'' also implies constraints on the object of adaptation, excluding a ``closed environment'' specific to a single or a set of particular problems, and considering there are no clearly predefined boundaries for the problems to be solved. With limited resources, faced with an open environment, the knowledge and resources of an intelligent agent are insufficient. 

This interpretation of ``intelligence'' (in Def.~\ref{def:intelligence}) takes into account current mainstream research (\textit{i.e.}, ML) and can be extended to future research (\textit{i.e.}, AGI). Based on this, the interpretation of ``general intelligence'' considers both the characteristics of the subject (\textit{i.e.}, responding to environmental changes) and clearly defines the boundaries of the objectives (\textit{i.e.}, non-specific problems).

If a system can adapt to the open environment, then it can learn to do a wide range of tasks. In contrast, a system capable of solving plenty of problems does not have to learn: programmers can tell it how to do directly or indirectly. 

\subsection{Defining ``Artificial General Intelligence''}

\begin{define}\label{def:agi}
An Artificial General Intelligence (AGI) system is a computer that is adaptive to the open environment with limited computational resources and that satisfies certain principles.
\end{define} The set of principles here is denoted as $\mathcal{P}_G$.

For AGI, problems are not predetermined and not specified ones, otherwise, there is most probably always a special system that performs better than any general system. Nevertheless, it should be noted that we can still seek a set of ``meta-problems'' to solve. For example, to predict future events might be an alternative meta-problem.



There are some other perspectives on defining AGI, and again, disagreements among them have been summarized by  \cite{goertzel2014agi} and \cite{wang2019defining}.
It is argued that the controversial part is on the set $\mathcal{P}_G$. 
Firstly, we have to do abstraction from observations when describing what intelligence looks like. Even in neuromorphic computing~\cite{gerstner2014snn}, it is impossible to simulate all the details of human brains. Some factors irrelevant to intelligence are somehow ignored -- it is not necessary to simulate molecular motion inside neurons and synapses in the context of AI.  As summarized by \cite{wang2019defining}, researchers see intelligence in different abstraction levels. 
Secondly, researchers with different backgrounds describe $\mathcal{P}_G$ with different scopes and formal languages. 
For example, from psychologists' and cognitive scientists' views, AGI should be characterized via cognitive functions, \textit{e.g.}, perception, reasoning, planning, motivation, emotion, \textit{etc}.  Principles on how those functions should be designed and realized might be contained in  $\mathcal{P}_G$. 
Some researchers, both in the cognitive architecture field and mainstream AI, listed a number of characteristics an AGI system should possess, \textit{e.g.}, being hierarchically structured, capability of incremental and online learning,  capability of symbolic reasoning, capability of causal inference, \textit{etc}. They are describing the principles corresponding to those characteristics or capabilities. 
Researchers focusing on problem-solving capabilities tend not to get inspiration from psychology and neuroscience directly but rather look for approaches that benefit performance. They might describe principles in $\mathcal{P}_G$ via the computer science language or argue that $\mathcal{P}_G$ is empty. 
For example, Morris et. al.~\cite{morris2023levels} call on focusing on problem-solving capabilities, and ignoring mechanisms, consequently implying  $\mathcal{P}_G$ to be empty. In neuroscience, the famous STDP (Spike-Timing-Dependent Plasticity)~\cite{gerstner2014snn} mechanism describes how the learning procedure should work. If one summarizes STDP into a principle, it might be that two successive events (represented by neuronal activation) tend to be associated with each other (via synaptic plasticity). At the most abstract level, it was argued that working in real-time is a necessary property for intelligence~\cite{wang2019defining}, and the corresponding principle should be included in $\mathcal{P}_G$. Researchers usually design their systems or methods in detail, however, some general principles should be finally extracted from their works, helping humanity to understand intelligence, especially how our mind works.

In this paper, it is impossible to argue what kind of descriptions or formal languages are better than others, nor what principles should be included in $\mathcal{P}_G$. Nonetheless, I hope readers could be aware of what the true controversial part is.

\section{Comparison to Other Definitions} \label{sec:comparisons}

Various definitions of intelligence and AGI were proposed.
In the ``sparks of AGI'' paper, 
``We use AGI to refer to systems that demonstrate broad capabilities of intelligence, including reasoning, planning, and the ability to learn from experience, and with these capabilities at or above human-level.''~\cite{bubeck2023sparks-agi}
This is certainly reasonable. According to this paper, the ability to learn is the most indispensable, while other features can be described in a disputable part.
A recent work on describing AGI is from DeepMind~\cite{morris2023levels} (``levels of AGI'') -- in a nutshell, they interpreted AGI as \textit{a computer that is capable of solving human-solvable problems, but not necessarily in human-like ways}. According to different breadths (\textit{i.e.}, range of tasks) and depths (\textit{i.e.}, performance on a task), they categorized AGI into different levels. The ``levels of AGI'' paper strongly emphasizes problem-solving capabilities but overlooks the process of acquiring these abilities, potentially leading research back to the old path of the ``AI effect''. Moreover, according to their definition, the best way to achieve AGI might be to design specific solutions for various problems and then integrate them on a computer, which, despite solving many problems very well, might not lead to a much deeper understanding of our own mind. Additionally, solving problems can be the result of \textit{adaptation}, and adaptability ultimately reflects in the performance of problem-solving. Compared to solving problems, stressing \textit{adaptation} in the definition is more instructive.
The Wang's definition of intelligence~\cite{wang2019defining} is almost equivalent to the definition of \textit{general intelligence} in this paper. To adapt to open environments with limited resources, a direct implication is that a system's knowledge and resources are insufficient, and this corresponds to the Assumption of Insufficient Knowledge and Resources (AIKR) in ~\cite{wang2019defining}. However, I consider the principles of intelligence, which are abstract descriptions of how AGI systems work, as a part of the definition, so that many previous definitions can be unified into a single one. In addition, the controversial part of Wang's definition can be put into the principles $\mathcal{P}_G$ (for example, not all researchers consider working in real-time as a key feature).

\section{Conclusion}

Some researchers refer intelligence to \textit{problem-solving capability} (\textit{e.g.}, works in the early stage of AI),  some refer intelligence to \textit{general intelligence} (e.g., works of AGI) in Def.~\ref{def:gi}, and some refer intelligence to the learning part in Def.~\ref{def:intelligence}. 
At the current stage, we have seen the huge success of Machine Learning, but that is not a reason for complacency. In contrast, with a better understanding and definition of (general) intelligence, we will readily embrace the age of AGI, realizing the original dream of the AI field. I believe that in the future, perhaps by the end of this century, when people talk about \textit{general intelligence}, the word ``general'' will not need to be explicitly emphasized to avoid misunderstanding.

The motivation of this paper is not to propose a fully novel definition of intelligence, but to review the previous, genius ideas and summarize a definition that can be used as a basic specification for the researchers both inside and outside the community.
In Def.~\ref{def:intelligence}, \ref{def:gi}, and \ref{def:agi}, I keep the part ``certain principles'' to be blurry, waiting for future discussions and debates on it. On the contrary, I make the part  ``adapt to open environments with limited resources'' stated clearly, so that two different concepts, \textit{intelligence} and \textit{problem-solving capability}, can be easily distinguished and it reminds researchers of the initial goal of the field -- to create thinking machines, whatever it is called.

%
%
\bibliographystyle{splncs04}
\bibliography{mybibliography}
%




\end{document}